\documentclass[11pt]{article}
\usepackage[margin=1in]{geometry}
\usepackage{setspace}
\usepackage{lmodern}
\usepackage[T1]{fontenc}
\usepackage{microtype}
\usepackage{amsmath, amssymb}
\usepackage{graphicx}
\usepackage{hyperref}
\hypersetup{
  colorlinks=true,
  linkcolor=blue,
  citecolor=blue,
  urlcolor=blue
}
\usepackage[numbers]{natbib}
\usepackage{titlesec}
\titleformat{\section}{\large\bfseries}{\thesection}{1em}{}
\titleformat{\subsection}{\normalsize\bfseries}{\thesubsection}{1em}{}
\title{\bfseries Improving Deep Knowledge Tracing via Gated Architectures and Adaptive Optimization}
\author{Altun Shukurlu \thanks{University of Virginia, \texttt{as4xa@virginia.edu}} }
\date{}

\begin{document}

\maketitle

\begin{abstract} Deep Knowledge Tracing (DKT)\cite{piech2015deep} models student learning behavior by leveraging Recurrent Neural Networks (RNNs) to predict future performance based on historical interaction data. However, the original implementation relied on standard RNNs in the Lua-based Torch framework, limiting extensibility and reproducibility. In this paper, we revisit the DKT model from two perspectives: architectural improvements and optimization efficiency. First, we enhance the model using gated recurrent units, specifically Long Short-Term Memory (LSTM)\cite{hochreiter1997long} and Gated Recurrent Units (GRU)\cite{cho2014learning}, which better capture long-term dependencies and mitigate vanishing gradient issues\cite{bengio1994learning, graves2013speech}. Second, we re-implement DKT using the PyTorch framework~\cite{paszke2019pytorch, 2017pytorch, yang2019pytorch}, enabling a modular and accessible infrastructure compatible with modern deep learning pipelines. Furthermore, we benchmark multiple optimization algorithms—SGD~\cite{robbins1951stochastic}, RMSProp~\cite{rmsprop}, Adagrad~\cite{duchi2011adaptive}, Adam~\cite{kingma2014adam}, and AdamW~\cite{loshchilov2019decoupled}—to evaluate their impact on convergence speed and predictive accuracy in educational modeling tasks. Experiments on the Synthetic-5 and Khan Academy datasets show that GRUs and LSTMs achieve higher accuracy and improved training stability over basic RNNs, while adaptive optimizers like Adam and AdamW outperform SGD in both early-stage learning and final model performance. Our open-source PyTorch implementation provides a reproducible and extensible foundation for future research in neural knowledge tracing and personalized learning systems. 
\end{abstract}

\section{Introduction}

Understanding how students acquire knowledge over time is a central challenge in educational data mining and intelligent tutoring systems. Accurate prediction of student performance enables adaptive learning platforms to provide timely feedback, support interventions, and personalize educational content. Traditional approaches to this task include probabilistic models such as \textit{Bayesian Knowledge Tracing (BKT)}\cite{corbett1994knowledge} and \textit{Performance Factor Analysis (PFA)}\cite{pavlik2009performance}, which represent student knowledge as a set of latent variables that evolve over time based on observed responses. While interpretable, these models rely on handcrafted features and often make simplifying assumptions about how learning unfolds.

\textit{Deep Knowledge Tracing (DKT)}\cite{piech2015deep} marked a major shift by applying recurrent neural networks (RNNs), specifically Long Short-Term Memory (LSTM) networks\cite{hochreiter1997long}, to model student interaction sequences directly. This approach eliminated the need for feature engineering, allowing the model to learn complex temporal patterns from raw input data. DKT has since become a strong baseline in student modeling research and inspired a growing body of work exploring attention mechanisms~\cite{pandey2019self, ghosh2020context}, memory networks~\cite{zhang2017dynamic}, graph structures~\cite{nakagawa2019graph}, and interpretability.

Despite its influence, the original DKT pipeline exhibits two key limitations that hinder reproducibility and modern usage:

\begin{enumerate} \item \textbf{Model architecture}: The original DKT model used standard RNNs, which are prone to vanishing gradients and struggle with long-term dependencies~\cite{bengio1994learning}. More advanced gated architectures—such as LSTMs and Gated Recurrent Units (GRUs)\cite{cho2014learning}—can better capture long-range temporal information via learned gating mechanisms that control memory updates over time\cite{graves2013speech}.

\item \textbf{Implementation infrastructure and optimization}: The original implementation used the Lua-based Torch framework, which is now deprecated and poorly supported. Additionally, the model was trained using vanilla Stochastic Gradient Descent (SGD)\cite{robbins1951stochastic}, without systematic evaluation of modern optimization algorithms. In contrast, frameworks like PyTorch\cite{paszke2019pytorch, 2017pytorch, yang2019pytorch} support high-level abstractions (e.g., \texttt{nn.LSTM}) and a wide range of adaptive optimizers, such as RMSProp~\cite{rmsprop}, Adagrad~\cite{duchi2011adaptive}, Adam~\cite{kingma2014adam}, and AdamW~\cite{loshchilov2019decoupled}, which offer improved convergence and generalization performance. \end{enumerate}

In this work, we revisit the DKT framework with the goal of modernizing its core components. Our contributions are threefold: \begin{itemize} \item We replace the original RNN with LSTM and GRU architectures to improve long-sequence modeling and training stability. \item We re-implement the model in PyTorch with a modular and extensible codebase that supports reproducible research and future development. \item We benchmark several optimization algorithms on the DKT task using the Synthetic-5 and Khan Academy datasets, showing that adaptive optimizers, particularly Adam and AdamW, outperform SGD in both convergence speed and predictive accuracy. \end{itemize}

Through these improvements, we aim to provide a clean, modern foundation for deep knowledge tracing research and encourage broader adoption of best practices in educational modeling pipelines. Our source code is publicly available at \url{https://github.com/Altunshukurlu/Modern_DeepKnowledgeTracing} and \url{https://github.com/Altunshukurlu/Optimized_DeepKnowledgeTracing}.

\section{Preliminaries}

To contextualize our improvements to the DKT framework, this section reviews the underlying components relevant to our work. We begin with an overview of Deep Knowledge Tracing and its evolution from traditional student modeling approaches. We then introduce recurrent neural networks (RNNs) and their gated variants—LSTM and GRU—which address key limitations of standard RNNs. Finally, we discuss modern optimization algorithms that influence training stability and convergence, setting the stage for our empirical evaluations.

\subsection{Deep Knowledge Tracing}

Knowledge Tracing (KT) is the task of modeling a student’s evolving knowledge state by analyzing their historical interactions with educational content~\cite{corbett1994knowledge}. The objective is to estimate the probability that a student will correctly answer future questions, which enables personalized feedback and adaptive learning systems. Traditional approaches like \textit{Bayesian Knowledge Tracing (BKT)}~\cite{corbett1994knowledge} and \textit{Performance Factor Analysis (PFA)}~\cite{pavlik2009performance} rely on interpretable probabilistic models but often require hand-crafted features and make strong assumptions about learning dynamics.

The introduction of \textit{Deep Knowledge Tracing (DKT)} by \citet{piech2015deep} marked a turning point. By applying recurrent neural networks to student interaction data, DKT allowed models to learn complex patterns directly from raw sequences, without the need for manual feature engineering. The success of DKT sparked a wave of research exploring ways to enhance its predictive power and interpretability. For instance, \textit{Dynamic Key-Value Memory Networks (DKVMN)}~\citep{zhang2017dynamic} introduced external memory to better represent concept-level knowledge, while \textit{Self-Attentive Knowledge Tracing (SAKT)}~\citep{pandey2019self} and \textit{Attentive Knowledge Tracing (AKT)}~\citep{ghosh2020context} replaced recurrence with attention mechanisms for more flexible modeling. Surveys such as those by \citet{abdelrahman2023knowledge} and \citet{liu2021survey} provide a comprehensive overview of this growing body of work, highlighting how deep learning has become integral to personalized learning systems.

\subsection{Recurrent Neural Networks}

\begin{figure}[h!]
\centering
\includegraphics[width=0.6\textwidth]{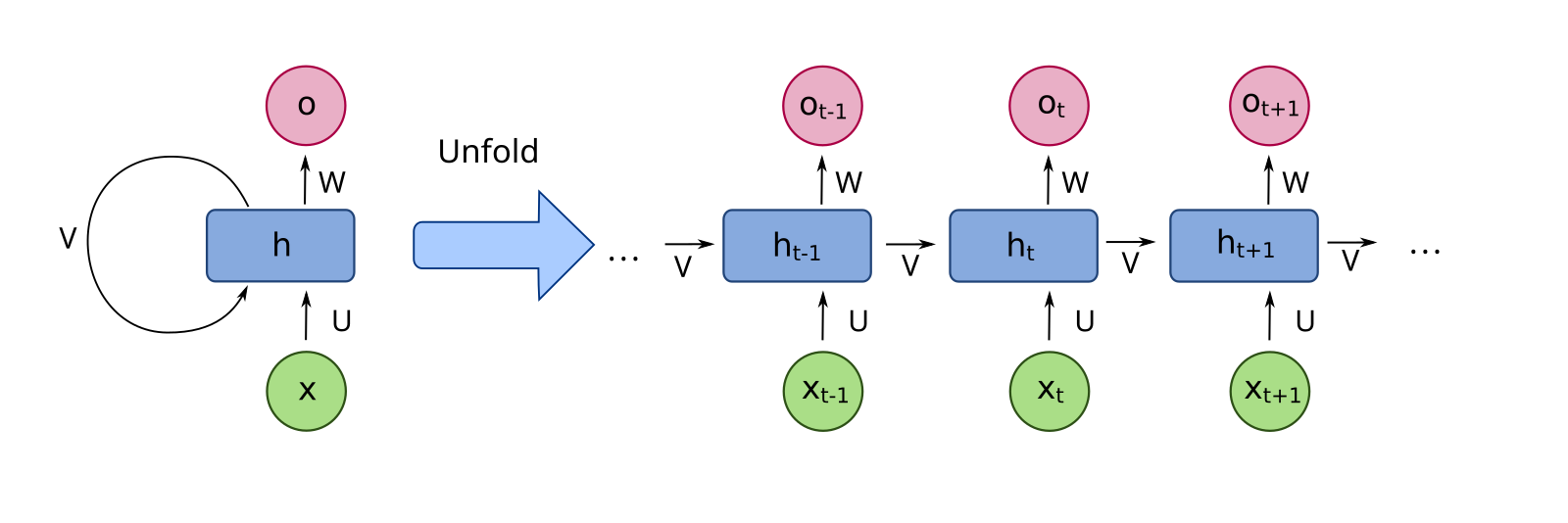}
\caption{Unfolded RNN architecture}
\end{figure}

Recurrent Neural Networks (RNNs) are a class of neural networks designed to process sequential data by maintaining a hidden state that evolves over time. A key feature of RNNs is the sharing of parameters across different time steps, enabling the model to capture temporal dependencies in the data.
Formally, a standard RNN is defined by the following recurrence relations:
\[
    h_t = \tanh(W_{xh}x_t + W_{hh}h_{t-1} + b_h)
\]
where \( h_t \) is the hidden state at time step \( t \), \( x_t \) is the input at time step \( t \), and \( W_{xh}, W_{hh} \) are the weight matrices connecting the input and previous hidden state to the current hidden state. \( b_h \) is the bias term.The output at each time step is computed as:
\[
    y_t = W_{hy}h_t + b_y
\]
where \( W_{hy} \) connects the hidden state to the output, and \( b_y \) is the output bias. This formulation allows RNNs to model sequences effectively; however, standard RNNs often struggle to capture long-range dependencies due to the vanishing gradient problem~\citep{bengio1994learning}.

To overcome these limitations, \citet{hochreiter1997long} proposed \textit{Long Short-Term Memory (LSTM)} networks, which incorporate gating mechanisms to regulate the flow of information over time. These gates allow the network to selectively retain or discard information at each time step, thereby enabling more stable and effective learning over extended sequences. In the context of Deep Knowledge Tracing, LSTMs have become a widely adopted backbone architecture~\citep{elman1990finding, sutskever2013training}, offering a robust and interpretable approach for modeling the progression of student knowledge. While newer alternatives such as attention mechanisms and graph-based models~\citep{nakagawa2019graph} have emerged, LSTMs continue to serve as a strong baseline in sequential educational modeling.

\subsection{LSTM and GRU Architectures}

To mitigate the limitations of standard RNNs, gated architectures such as Long Short-Term Memory (LSTM) networks~\cite{hochreiter1997long} and Gated Recurrent Units (GRUs)~\cite{cho2014learning} introduce gating mechanisms that regulate the flow of information over time.

\begin{figure}[h!]
\centering
\includegraphics[width=0.6\textwidth]{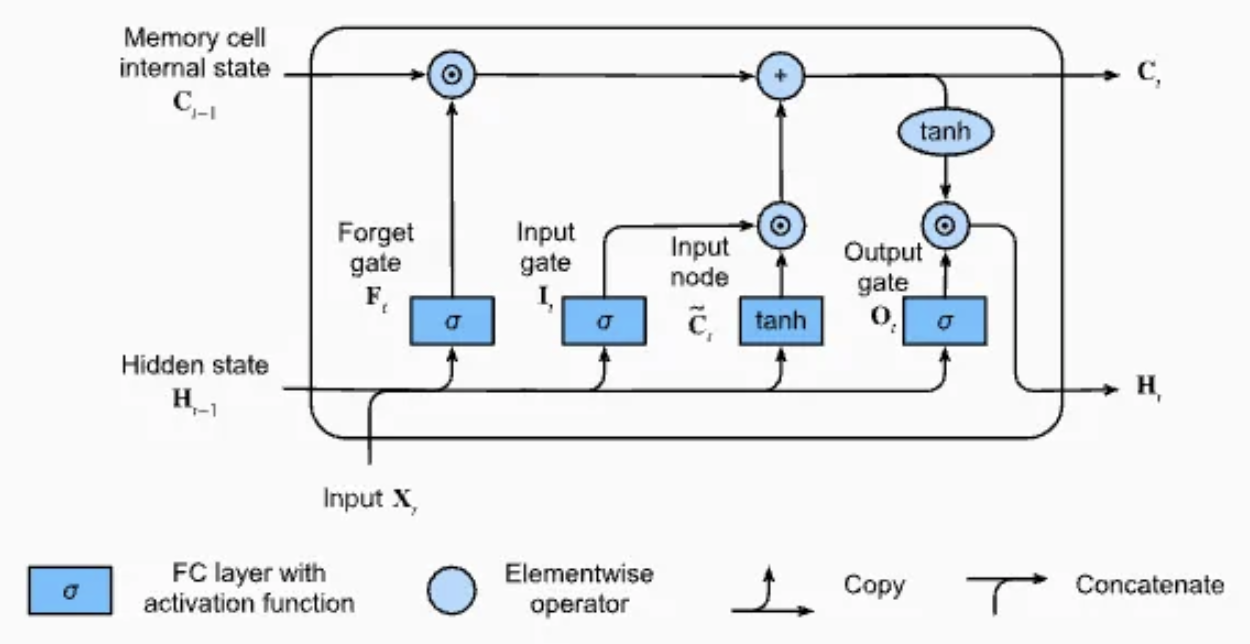}
\caption{Gating Mechanisms in LSTM architecture}
\end{figure}

The core components of an LSTM are the \textit{forget gate}, \textit{input gate}, and \textit{output gate}, each of which controls how information is passed through the network. The gating mechanisms are formulated as follows:

\[
\begin{aligned}
    f_t &= \sigma(W_f x_t + U_f h_{t-1} + b_f) \\
    i_t &= \sigma(W_i x_t + U_i h_{t-1} + b_i) \\
    o_t &= \sigma(W_o x_t + U_o h_{t-1} + b_o) \\
    \tilde{c}_t &= \tanh(W_c x_t + U_c h_{t-1} + b_c) \\
    c_t &= f_t \odot c_{t-1} + i_t \odot \tilde{c}_t \\
    h_t &= o_t \odot \tanh(c_t)
\end{aligned}
\]

where \( f_t \) is the forget gate, controlling the amount of previous memory to forget, \( i_t \) is the input gate, determining how much of the current input is used to update the memory and \( \tilde{C}_t \) represents the candidate memory cell, which is generated from the input. Moreover, \( C_t \) is the cell state, which is updated at each time step based on the forget and input gates and  \( o_t \) is the output gate, regulating the amount of information passed to the hidden state \( h_t \).
These equations allow the LSTM to selectively remember or forget information, making it particularly effective for tasks where long-term dependencies are crucial, such as natural language processing \cite{bengio1994learning} and time series prediction \cite{graves2013speech}.

Gated Recurrent Units (GRUs) \cite{cho2014learning} were introduced as a more computationally efficient alternative to Long Short-Term Memory (LSTM) networks. GRUs aim to address the same problem of capturing long-term dependencies in sequences but with a simpler architecture that combines the forget and input gates into a single update gate. This reduction in complexity often results in faster training and reduced computational resources while still maintaining strong performance in tasks requiring sequential data modeling.
\[
\begin{aligned}
    z_t &= \sigma(W_z x_t + U_z h_{t-1} + b_z) \\
    r_t &= \sigma(W_r x_t + U_r h_{t-1} + b_r) \\
    \tilde{h}_t &= \tanh(W_h x_t + U_h (r_t \odot h_{t-1}) + b_h) \\
    h_t &= (1 - z_t) \odot h_{t-1} + z_t \odot \tilde{h}_t
\end{aligned}
\]
The core components of a GRU are the \textit{update gate} and the \textit{reset gate}, which control the information.

\section{Modern Optimization Algorithms on DKT}

We evaluate the performance of several optimization algorithms on Deep Knowledge Tracing (DKT) with LSTM models in a single epoch. Each optimizer defines a distinct strategy for updating the network parameters $\theta$ using the gradients $\nabla_\theta \mathcal{L}$ of the loss function. Below, we describe each optimizer formally, very briefly:

\textbf{Stochastic Gradient Descent (SGD)}
Stochastic Gradient Descent updates parameters using noisy estimates of the gradient based on mini-batches:
\begin{equation}
    \theta_{t+1} = \theta_t - \eta \nabla_\theta \mathcal{L}(\theta_t),
\end{equation}
where $\eta$ is the learning rate. While simple and computationally efficient, vanilla SGD is sensitive to the choice of $\eta$ and often suffers from slow convergence, especially in ravine-like loss landscapes.

\textbf{RMSProp} adapts the learning rate for each parameter based on a decaying average of squared gradients:
\begin{align*}
    s_{t+1} &= \rho s_t + (1 - \rho)(\nabla_\theta \mathcal{L}(\theta_t))^2 \\
    \theta_{t+1} &= \theta_t - \frac{\eta}{\sqrt{s_{t+1} + \epsilon}} \nabla_\theta \mathcal{L}(\theta_t),
\end{align*}
where $s_t$ is the moving average of squared gradients, $\rho$ is the decay rate (commonly 0.9), and $\epsilon$ is a small constant for numerical stability. RMSProp is particularly effective in handling non-stationary objectives and ill-conditioned curvature.

\textbf{Adam} ~\citep{kingma2014adam} combines the benefits of Momentum and RMSProp by maintaining exponentially decaying averages of both first and second moments of the gradients:
\begin{align*}
    m_{t+1} &= \beta_1 m_t + (1 - \beta_1)\nabla_\theta \mathcal{L}(\theta_t) \\
    v_{t+1} &= \beta_2 v_t + (1 - \beta_2)(\nabla_\theta \mathcal{L}(\theta_t))^2 \\
    \hat{m}_{t+1} &= \frac{m_{t+1}}{1 - \beta_1^{t+1}}, \quad 
    \hat{v}_{t+1} = \frac{v_{t+1}}{1 - \beta_2^{t+1}} \\
    \theta_{t+1} &= \theta_t - \eta \frac{\hat{m}_{t+1}}{\sqrt{\hat{v}_{t+1}} + \epsilon},
\end{align*}
where $\beta_1, \beta_2$ are decay rates for the first and second moment estimates (typically $\beta_1 = 0.9$, $\beta_2 = 0.999$). Adam is widely used due to its fast convergence and robustness to hyperparameter tuning.

\textbf{Adagrad} adapts the learning rate to each parameter based on the historical sum of squared gradients:
\begin{align}
    G_{t+1} &= G_t + (\nabla_\theta \mathcal{L}(\theta_t))^2 \\
    \theta_{t+1} &= \theta_t - \frac{\eta}{\sqrt{G_{t+1}} + \epsilon} \nabla_\theta \mathcal{L}(\theta_t),
\end{align}
where $G_t$ is a diagonal matrix containing the accumulated squared gradients. Adagrad is especially useful in settings with sparse features, but its learning rate may decay too aggressively over time.

\bigskip

In our experiments, we evaluate each optimizer on the same LSTM-based DKT architecture and dataset, comparing their convergence speed and final predictive performance. Results are presented in Section~\ref{sec:results}.

\section{Experiment Results}
\label{sec:results}

In this section, we evaluate the performance of our modernized Deep Knowledge Tracing (DKT) models along two primary dimensions: (1) the effect of gated recurrent architectures (LSTM vs GRU), and (2) the impact of optimization strategies on training efficiency and predictive accuracy. Experiments were conducted on the Synthetic-5 dataset~\cite{piech2015deep} and a large-scale subset of the Khan Academy dataset.

\subsection{LSTM vs GRU: Architecture Comparison}

We evaluated our modernized implementation on the Synthetic-5 dataset~\cite{piech2015deep}, which was also used in the original DKT paper. Our experiments included both LSTM and GRU-based models. During training, we utilized the binary cross-entropy loss function with logits, defined as:

\[
\texttt{self.criterion = nn.BCEWithLogitsLoss(reduction='none')}
\]

This loss function was computed throughout the training process, and model performance was assessed using both AUC (Area Under the Curve) and prediction accuracy metrics. The LSTM-based model demonstrated comparable or superior AUC scores and accuracy relative to the GRU model, while also exhibiting improved training stability compared to the original RNN-based approach.

Further qualitative analysis revealed that the LSTM model excelled in handling longer sequences, effectively minimizing the prediction oscillations that were prevalent in the original RNN-based implementation. Although the GRU model performed competitively, with faster computational efficiency, the LSTM model outperformed it in terms of both stability and predictive accuracy.

\begin{table}[ht]
\centering
\begin{tabular}{|l|l|c|c|c|c|}
\hline
 \textbf{Model} & \textbf{Train Loss} & \textbf{Train Accuracy} & \textbf{Test Accuracy} & \textbf{Time (s)} \\ \hline
 LSTM & 0.6166 & 0.6344 & 0.6356 & 387.1 \\ \hline
 GRU  & 0.6522 & 0.6344 & 0.6356 & 352.7 \\ \hline
\end{tabular}
\caption{Performance Comparison: LSTM vs GRU}
\end{table}

\subsection{Optimizer Benchmarking on Khan Academy}

We present a detailed numerical analysis of the performance of various optimization algorithms in the context of training Deep Knowledge Tracing (DKT) models on the Khan Academy dataset. The primary objective of this analysis is to evaluate the effectiveness of different optimizers in terms of both convergence speed and predictive accuracy, particularly during the early stages of training.

\subsubsection{Convergence Performance After 1 Epoch}

Given the large scale of the dataset and the computational costs associated with prolonged training, we focus on performance metrics from the first epoch to provide insights into how efficiently each optimizer can adapt to the data and begin to minimize the training error.

\begin{table}[h]
\centering
\begin{tabular}{|l|c|c|}
\hline
\textbf{Optimizer} & \textbf{Error} & \textbf{Accuracy} \\
\hline
SGD     & 0.6946 & 0.4513 \\
RMSProp & 0.6598 & 0.5542 \\
Adagrad & 0.6557 & 0.5546 \\
Adam    & 0.6548 & 0.5384 \\
AdamW   & 0.6549 & \textbf{0.5550} \\
\hline
\end{tabular}
\caption{Convergence performance after a single training epoch. Adaptive optimizers show significantly stronger early-stage accuracy.}
\label{tab:epoch1-convergence}
\end{table}

Table~\ref{tab:epoch1-convergence} summarizes the convergence performance after the first epoch of training. It is evident that adaptive optimizers (Adam, AdamW, RMSProp, and Adagrad) are notably more effective in reducing training error and achieving higher accuracy compared to the stochastic gradient descent (SGD) optimizer. Specifically, all adaptive methods reach an accuracy of over 53\%, whereas SGD only achieves 45\%. Among the adaptive optimizers, AdamW slightly outperforms the others, achieving the highest accuracy after the first epoch. This result suggests that adaptive methods are not only more efficient in early-stage convergence but also offer better responsiveness to the data.

\subsubsection{Training Time Comparison}

To evaluate the computational trade-offs of different optimizers, we measured training time per epoch under identical conditions.

\begin{table}[h]
\centering
\begin{tabular}{|l|c|}
\hline
\textbf{Optimizer} & \textbf{Per Epoch Training Time (s)} \\
\hline
SGD     & 764.34 \\
RMSProp & 801.53 \\
Adagrad & 809.50 \\
Adam    & 778.64 \\
AdamW   & 776.55 \\
\hline
\end{tabular}
\caption{Training time per epoch for each optimizer. Differences are minor but noteworthy for large-scale training scenarios.}
\label{tab:epoch1-speed}
\end{table}

The training times for each optimizer, as reported in Table~\ref{tab:epoch1-speed}, exhibit minor differences, with adaptive optimizers like Adagrad and RMSProp incurring a slight overhead (approximately 30–40 seconds more per epoch). However, these differences are relatively insignificant when considering the substantial improvements in convergence speed and early-stage accuracy. This indicates that, for large-scale training scenarios, the benefits derived from using adaptive optimization strategies outweigh the modest increase in training time.

\subsubsection{Key Insights}

These findings underscore the importance of selecting appropriate optimization techniques in the early stages of training, particularly when working with large datasets. Adaptive optimizers provide substantial advantages in terms of both speed and accuracy, making them highly advantageous for scenarios where rapid performance feedback is necessary, such as in real-time educational systems.

\section{Conclusion}

In this work, we revisited the Deep Knowledge Tracing (DKT) framework with the goal of modernizing its model architecture and training pipeline. First, we replaced the original RNN backbone with gated recurrent architectures—Long Short-Term Memory (LSTM) networks and Gated Recurrent Units (GRUs)—which improved the model's ability to capture long-range dependencies and stabilized training. Second, we migrated the implementation from the deprecated Lua-based Torch framework to PyTorch, enabling a modular, extensible, and reproducible infrastructure aligned with current deep learning practices.

We also conducted a comprehensive benchmark of optimization strategies and found that adaptive optimizers—particularly Adam and AdamW—outperform vanilla SGD in terms of convergence speed, training stability, and predictive accuracy. AdamW, in particular, provided the best trade-off between performance and efficiency.

Beyond methodological improvements, our findings have practical implications for real-world educational systems. GRUs offer comparable predictive accuracy to LSTMs with faster training times, making them well-suited for time-sensitive or resource-constrained environments. The rapid convergence of AdamW further supports its use in online learning settings, where fast model updates are essential. By open-sourcing a clean and extensible PyTorch implementation, we provide a foundation for future development in personalized learning technologies.

Future work will explore newer optimizers, such as AdaBelief and Lookahead, incorporate attention mechanisms, and expand benchmarking across diverse educational datasets. We also aim to improve the interpretability of learned knowledge representations and visualize student learning trajectories to support meaningful pedagogical insights.

\bibliography{main}
\end{document}